\documentclass[10pt,twocolumn,letterpaper]{article}
\usepackage{graphicx}
\usepackage{amsmath}
\usepackage{amssymb}
\usepackage{booktabs}
\usepackage{multirow}
\usepackage{caption}
\usepackage{algpseudocode}
\usepackage{subcaption}
\usepackage[table]{xcolor}
\usepackage{multirow}
\usepackage{amsmath}
\usepackage{amssymb}
\usepackage{setspace} 
\usepackage{lipsum}
\usepackage[margin=1.5cm]{geometry}
\usepackage{algorithm}
\usepackage{kotex} 
\usepackage[dvipsnames]{xcolor}
\usepackage{algpseudocode}
\usepackage{amssymb}
\usepackage{pifont}
\usepackage{amssymb}
\usepackage{amsfonts}
\usepackage{gradient-text}
\definecolor{imgcolor}{HTML}{2684F7}
\definecolor{vidcolor}{HTML}{F03C4E}
\usepackage{xcolor} 
\usepackage{colortbl} 
\usepackage{array} 
\usepackage{dsfont}
\usepackage[table]{xcolor}

\newcommand{\huggingface}{\raisebox{-1.5pt}{\includegraphics[height=1.05em]{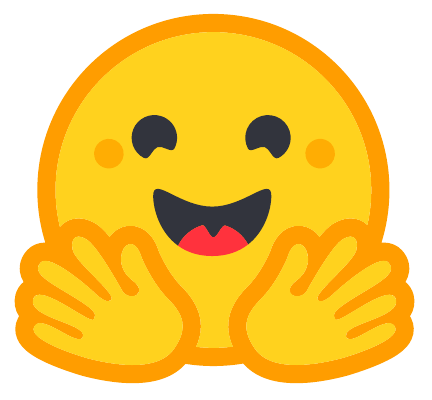}}\xspace}
\newcommand{\github}{\raisebox{-1.5pt}{\includegraphics[height=1.05em]{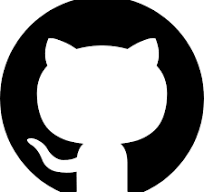}}\xspace}
\renewcommand{\thefootnote}{\fnsymbol{footnote}}

\usepackage[pagenumbers]{wacv} 

\definecolor{wacvblue}{rgb}{0.21,0.49,0.74}
\usepackage[pagebackref,breaklinks,colorlinks,allcolors=wacvblue]{hyperref}

\def\wacvPaperID{104} 
\def\confName{WACV}
\def\confYear{2026}

\begin{document}

\title{ForestSplats: Deformable transient field for Gaussian Splatting in the Wild}

\author{
  Wongi Park$^{1}$\thanks{Equal contribution.}, Myeongseok Nam$^{1}$\footnotemark[1], Siwon Kim$^1$, Sangwoo Jo$^2$ and Soomok Lee$^1$\thanks{Corresponding author.}\footnotemark[2] \\
  {$^{1}$Ajou University \;\;\; $^{2}$Minds and Company}  \\
  \texttt{\{psboys, soomoklee\}@ajou.ac.kr}
}

\twocolumn[{%
\renewcommand\twocolumn[1][]{#1}%
\maketitle
\begin{center}
    \vspace{-10mm}
    {\textbf{\textcolor{magenta}{\url{https://forestsplats.github.io/}}}}
\end{center}
\begin{center}
    \centering
    \captionsetup{type=figure}
    \includegraphics[width=\textwidth, height=5.0cm]{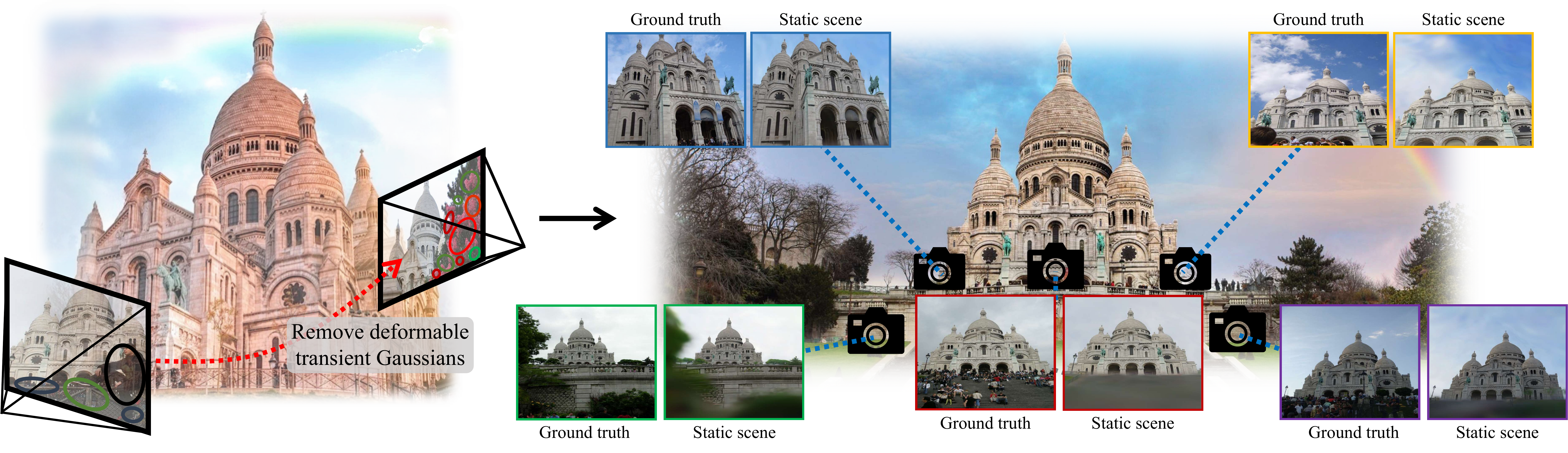}
    \caption{Given unconstrained images, ForestSplats efficiently represents transient elements from 2D scenes and achieves fast rendering speed while preserving high-quality. Our method decomposes the scene into the deformable transient field and the static field to disentangle static and transient elements without a Vision Foundation Model (VFM). More details are available in the supplementary materials.} 
\end{center}
}]
\let\thefootnote\relax
\footnotetext{
\github \textbf{Page}: \href{https://forestsplats.github.io/}{pull-ups/ForestSplats} \;\;
  \huggingface \textbf{Model}: \href{https://forestsplats.github.io/}{pull-ups/Collections}}
  
\begin{abstract}
\hspace{0.5em} Recently, 3D Gaussian Splatting (3D-GS) has emerged, showing real-time rendering speeds and high-quality results in static scenes. Although several 3D-GS methods show effectiveness in static scenes, their performance significantly degrades in real-world environments due to transient objects, lighting variations, and diverse levels of occlusion. To tackle this, existing methods estimate occluders or transient elements by leveraging pre-trained models or integrating additional transient field pipelines. However, these methods still suffer from two defects: \textbf{1)} Using semantic features from the Vision Foundation Model (VFM), such as DINO, might limit generalization on unseen data due to reliance on prior knowledge. \textbf{2)} The transient field requires significant memory to handle transient elements with per-view Gaussians and struggles to define clear boundaries for occluders, solely relying on photometric errors. To address these problems, we propose \textbf{ForestSplats}, a novel approach that leverages the deformable transient field and a superpixel-aware mask to efficiently represent transient elements in the 2D scene across unconstrained image collections and effectively decompose static scenes from transient distractors without VFM. We designed the transient field to be deformable, capturing per-view transient elements. Furthermore, we introduce a superpixel-aware mask that clearly defines the boundaries of occluders by considering photometric errors and superpixels. Additionally, we propose uncertainty-aware densification to avoid generating Gaussians within the boundaries of occluders during densification. Through extensive experiments across several benchmark datasets, we demonstrate that ForestSplats outperforms existing methods without VFM and shows significant memory efficiency in representing transient elements. \end{abstract}    
\section{Introduction}
\label{sec:intro}
\hspace{0.5em} Novel view synthesis from a sparse set of captured 2D images is a challenging and fundamental problem in computer vision and graphics, which has recently gained popularity. Such advancements play a critical role in applications such as augmented/virtual reality (VR) \cite{jiang2024vr, jiang2024robust}, autonomous driving \cite{li2024memorize, zhou2024drivinggaussian}, and 3D content generation \cite{liu2024one, xia2024video2game}. Even though Neural Radiance Field (NeRF) \cite{mildenhall2021nerf} has shown impressive results and rendering efficiency, it suffers from slow rendering speeds. Due to this issue, 3D Gaussian Splatting (3D-GS) \cite{kerbl20233d} recently has emerged as an explicit 3D representation that offers real-time rendering speeds. However, existing methods \cite{wu20244d, wang2023f2} assume that 2D images are captured without transient elements, various levels of occluders, or appearance variations such as changing sky, weather, and illumination. Therefore, accurately and robustly reconstructing with unconstrained image collections remains a primary challenge due to many transient elements in real-world environments. \\ 
\indent Several works have proposed various solutions to tackle this challenge. Early NeRF-based methods \cite{martin2021nerf, chen2022hallucinated, yang2023emernerf} introduce per-image transient embedding or photometric errors to typically ignore transient elements from unconstrained images. CR-NeRF \cite{yang2023cross} proposes a cross-ray paradigm and grid sampling strategy for efficient appearance modeling. Although showing promising results, these methods \cite{martin2021nerf,chen2022hallucinated,yang2023cross} are computationally intensive and struggle with real-time rendering. In contrast, 3D-GS-based methods \cite{xu2024wild, kulhanek2024wildgaussians, sabour2024spotlesssplats, lin2024hybridgs, wang2024desplat} usually utilize semantic features from Vision Foundation Model (VFM) to estimate precise distractors mask, showing fast rendering speed. WildGaussians \cite{kulhanek2024wildgaussians} also utilizes semantic features from DINOv2 \cite{oquab2023dinov2} to handle occlusions. HybridGS \cite{lin2024hybridgs} defines a transient field to explicitly decompose transient elements. Despite significant advancements, these methods still suffer from two limitations: \textbf{Firstly,} leveraging semantic features from the Vision Foundation Model (VFM) rely on prior knowledge, which restricts effective generalization to unseen data. \textbf{Secondly,} the transient field uses significant memory to handle transient elements using Gaussians per image and struggles to clearly estimate boundaries of occluders due to solely relying on the photometric errors. \\
\indent Our motivation stems from the lack of appearance variation and multi-view consistency in transient elements. That is, we only need to consider transient elements in the 2D scene from each view of unconstrained image scenarios. To end, we propose ForestSplats, which leverages a deformable transient field and superpixel-aware mask to effectively decouple transient elements from static scenes, showing high-quality rendering. We design deformable transient field that efficiently represents transient elements in the 2D scene across unconstrained image collections, which are explicitly decomposed from static field and show memory efficiency. Furthermore, we introduce superpixel-aware masking, which effectively separates distractors from static scenes without a pre-trained VFM by considering the photometric errors and superpixels in the 2D scene. Additionally, we incorporate a multi-stage training scheme to enhance the superpixel-aware mask in accurately capturing transient elements, achieving distractor-free novel-view synthesis. Furthermore, we introduce uncertainty-aware densification for static field, thereby avoiding generating Gaussians within the boundaries of occluders, showing high-quality rendering. Through extensive experiments on several datasets, we demonstrate that ForestSplats effectively decomposes transient elements from static scenes, outperforming the state-of-the-art methods  without VFM in novel view synthesis. Our primary contributions can be summarized as follows. \begin{itemize}
\item
We propose ForestSplats, which explicitly decomposes 3D scenes into static and deformable transient fields, effectively capturing static scenes while demonstrating memory efficiency to represent transient elements.
\item 
We introduce a superpixel-aware mask and a multi-stage training scheme that effectively separate the 3D scene representation into static and transient fields by considering the photometric errors and the superpixels.
\item
We introduce uncertainty-aware densification to avoid generating Gaussians within the boundaries of occluders.
\item
Extensive experiments demonstrate that ForestSplats achieves state-of-the-art performance and shows memory efficiency to represent transient elements.
\end{itemize}
\begin{figure*}[!th]
\centering
\includegraphics[width=\textwidth]{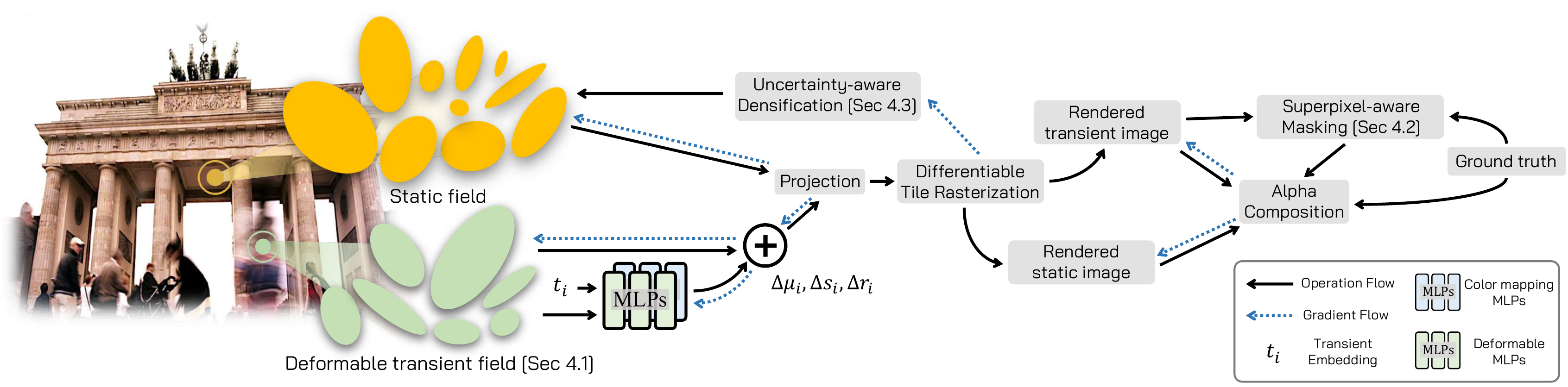}
\vspace{-2em}
\caption{An overview of ForestSplats. We decompose the 3D scene into the static and deformable transient fields. The rendered images combine alpha blending with a mask to produce compositional rendering results. Furthermore, we leverage a superpixel-aware mask, which considers photometric errors and superpixels to effectively decompose static and transient elements. Specifically, during optimization, we leverage uncertainty-aware densification (UAD) to avoid generating static Gaussians within the boundaries of occluders.}

\label{fig:overview}
\end{figure*}
\section{Related Works}
\label{sec:related}
\paragraph{3D Scene Decomposition.} Early decomposition-based approaches \cite{yang2023emernerf, nguyen2024rodus, ost2021neural} have been widely used in dynamic 3D novel view synthesis, decoupling dynamic objects from static scenes by comparing successive video frames and identifying differences as foreground elements. These methods \cite{yang2023emernerf, nguyen2024rodus, ost2021neural, turki2023suds} achieve more effective partitioning of static and dynamic elements by modeling scene dynamics through deformations of a canonical volume, particle-level motion, or rigid transformations of local geometric primitives. Meanwhile, another line of work \cite{martin2021nerf, li2023nerf, wang2024desplat, lin2024hybridgs} focuses on static scenes by disentangling transient objects and lighting variations while utilizing per-view transient Gaussians or a transient field. However, these approaches \cite{wang2024desplat, lin2024hybridgs} assign each Gaussian primitive to an individual image, increasing the number of transient Gaussian primitives, which requires inefficient memory usage. In contrast, we introduce a deformable transient field, enabling the efficient representation of transient elements across unconstrained image collections, showing high-quality rendering without excessive memory consumption.
\vspace{0.5\baselineskip}
\newline \textbf{Uncertainty in 3D Scene Reconstruction.} Recent novel view synthesis methods \cite{mildenhall2021nerf, kerbl20233d, shin2025locality} rely on the assumption of distractor-free environments. However, unconstrained image collections often violate these assumptions due to dynamic elements, lighting variations, and other transient factors. Numerous earlier methods \cite{martin2021nerf, mildenhall2021nerf,yang2023cross, li2023nerf} utilize implicit neural representation to encode 3D scenes, with differentiable volume rendering along the rays. NeRF-W \cite{martin2021nerf} optimizes a transient field and an uncertainty embedding per view to address distractors. CR-NeRF \cite{yang2023cross} introduces a cross-ray paradigm and a grid sampling strategy for efficient appearance modeling. NeRF-MS \cite{li2023nerf} tackles multi-sequence images captured in the wild using triplet loss and a decomposition module with several transient embeddings. While NeRF-based methods \cite{martin2021nerf, mildenhall2021nerf,yang2023cross, li2023nerf} offer efficient and high-quality 3D rendering, they are still computationally resource-intensive and limit real-time rendering speed. On the other hand, 3D-GS-based methods \cite{kerbl20233d, yang2024depth, xu2024wild, kulhanek2024wildgaussians} explicitly represent the 3D scene with differentiable rasterization, showing fast rendering speed and high-quality results. Wild-GS \cite{xu2024wild} leverages depth information from pre-trained Depth Anything \cite{yang2024depth} and a hierarchical decomposition strategy to regularize transient elements and preserve geometry consistency. WildGaussians \cite{kulhanek2024wildgaussians} also utilizes semantic features of pre-trained DINOv2 \cite{oquab2023dinov2} to handle transient elements. Unlike these approaches, our method effectively decomposes distractors from static scenes and represents transient elements without pre-trained VFM by leveraging photometric errors and superpixels.
\vspace{0.5\baselineskip}
\newline \textbf{Adaptive Density Control of 3D Gaussians.} Adaptive density control (ADC) strategy in 3D Gaussian Splatting works \cite{kerbl20233d} through two operations, densification and pruning, which capture empty areas or fine details during optimization. While showing efficiency, this ADC module relies on predetermined thresholds and does not leverage context information, leading to blurriness in high-frequency details in sparse point areas. To tackle this, Pixel-GS \cite{zhang2024pixel} leverages the number of pixels covered by Gaussian primitives as weights to promote densification, complementing average 2D position gradients. Taming-3DGS \cite{mallick2024taming} proposes score-based densification and steerable densification strategies, ordered by high opacity, improving the quality per Gaussian. Unfortunately, most existing methods \cite{kerbl20233d, zhang2024pixel, mallick2024taming} improve rendering quality in constrained image collections, but their application to unconstrained image collections remains challenging due to transient elements. In this work, we aim to develop an efficient densification strategy for casually captured images in the wild. To this end, we assign learnable uncertainty attributes to each Gaussian, removing high-uncertainty Gaussians in the static field. Furthermore, we effectively refine densification to avoid generating Gaussians within the boundaries of occluders during densification by considering positional gradient and the number of pixels covered by Gaussians. 

\section{Preliminary}
\paragraph{3D Gaussian Splatting (3D-GS).} 3D Gaussian Splatting (3D-GS) \cite{kerbl20233d} has emerged as a promising method for novel view synthesis, representing a scene as a set of differentiable Gaussians $\{ \mathcal{G}_i \}$. Each Gaussian $\mathcal{G}_i$ is parameterized by a position $ \mu_{i} \in \mathbb{R}^{3}$, covariance matrix $ \Sigma_{i} \in \mathbb{R}^{3 \times 3}$, which is decomposed into rotation matrix $R \in \text{SO(3)}$, and scaling matrix $S \in \mathbb{R}^{3 \times 3}$, an opacity $\alpha_{i} \in [0, 1]$, and view-dependent colors $c_{i} \in \mathcal{C}^{N_{sh}}$ represented via spherical harmonics (SH) coefficients $N_{sh}$. To render images, 3D Gaussians $\{ \mathcal{G}_i \}$ are projected onto screen space as 2D Gaussians via the viewing transformation $W \in \mathbb{R}^{3 \times 3}$ and covariance matrix $\Sigma'_{i} = JW \Sigma_{i} W^{T} J^{T} $ where $J \in \mathbb{R}^{2 \times 3}$ is the Jacobian of the Taylor approximation of the projective transformation. For each pixel, the $K$ Gaussians $\{ \mathcal{G}_k \; \vert \; k = 1, ..., K \}$ are ordered by depth and rendered using alpha-blending, resulting in pixel color $\hat{C}$: 
{\footnotesize\begin{equation}
\hat{C}=\sum\limits_{k=1}^{K} c_{k} \alpha_{k}  \prod_{j=1}^{k-1}\left(1-\alpha_j\right).\end{equation}}All attributes of Gaussian $\{ \mathcal{G}_i \}$ are optimized by minimizing the following reconstruction loss between rendered image $\hat{I} \in \mathbb{R}^{3 \times H \times W}$ and ground truth image
$I_{gt}  \in \mathbb{R}^{3 \times H \times W}$. \begin{equation}
\mathcal{L}_{\mathrm{GS}}=(1-\lambda) \mathcal{L}_1(\hat{I}, I_{gt})+\lambda \mathcal{L}_{\text {D-SSIM}}(\hat{I}, I_{gt} ), \end{equation}
where $\mathcal{L}_{1}$ is an $L_{1}$ loss, $\mathcal{L}_{\text{D-SSIM}}$ is a SSIM loss and $\lambda$ is a weighting factor. All 3D Gaussians are initialized using a sparse point
cloud obtained from the Structure-from-Motion (SfM) approach such as COLMAP \cite{schoenberger2016mvs}. Meanwhile, to effectively control the density of Gaussians and address under and over-reconstruction problems, adaptive densification and pruning strategies are employed to enhance the 3D scene reconstruction during optimization.
\begin{figure}[!t]
\centering
\includegraphics[width=\linewidth]{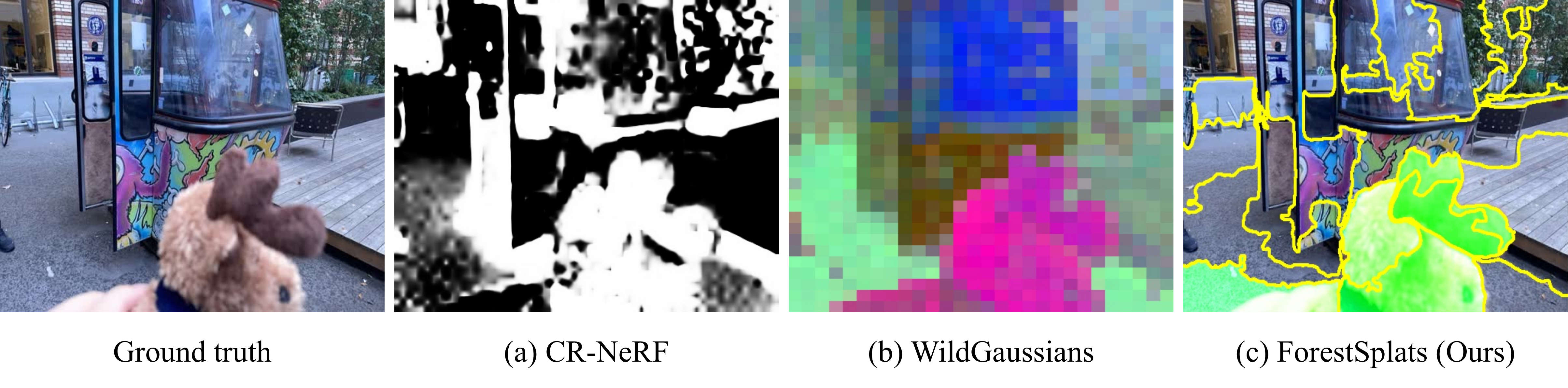}
\vspace{-2em}
\caption{Comparison of masking methods. (a) CR-NeRF focuses only on the photometric errors such as L1 loss, (b) WildGaussians considers the semantic features, relying on VFM, and (c) ForestSplats leverages photometric errors and superpixels to capture distractors, creating sharp boundaries of transients without VFM.}
\label{fig:method-mask}
\end{figure}

\section{Method}
\hspace{1em} Given unconstrained image collections, our goal is to represent the static scene and effectively remove all transient elements in 2D scenes. To this end, we propose ForestSplats, which leverages a deformable transient field and a superpixel-aware mask to effectively separate static scenes from transient distractors in 2D scenes. Firstly, we describe a deformable transient field to represent transient elements across unconstrained image collections without excessive memory usage in Sec. \ref{sec:4.1}. Then, in Sec. \ref{sec:4-2}, we propose a superpixel-aware mask to clearly separate a static scene from diverse occluders in 2D scenes. Finally, we introduce uncertainty-aware densification to avoid generating Gaussians within the boundaries of occluders, as described in Sec. \ref{sec:4-3}. The overview of our method is depicted in Fig. \ref{fig:overview}.
\subsection{Deformable Transient Field}
\label{sec:4.1}
\hspace{1em} Our primary objective is to decompose a 3D scene into transient and static fields while effectively identifying distractors. Although existing methods \cite{martin2021nerf,wang2024desplat, lin2024hybridgs, li2023nerf} utilize stationary transient field or per-view Gaussians to represent transient elements, prior methods are either limited in effectively capturing transient elements per image or show excessive memory usage, as shown in Tab.~\ref{tab:ablMem}. To mitigate these limitations, we propose a deformable transient field $\mathcal{G}_d$ to efficiently represent transient elements per image across unconstrained photo collections. We follow the paradigm of Deform3D-GS \cite{yang2024deformable} to represent transient elements for each view $I_{gt}$. Our method represents each view $\hat{I}$ as the sum of transients $\hat{I}_d \in \mathbb{R}^{3\times H\times W}$ and statics $\hat{I}_s \in \mathbb{R}^{3\times H\times W}$ through a mask $M \in [0, 1]$ such that: 
{\small\begin{equation}
    \hat{I} = M \odot \hat{I}_d+\left(1-M\right) \odot \hat{I}_s.
\end{equation}}Given transient embedding per image $t_i$ and position $\mu_i$ of transient 3D Gaussians $\mathcal{G}_{d}(\mu_i, s_i, r_i)$ as inputs, the deformation transient MLP $f_d$ predicts transient elements, producing the offset $\Delta$$\mu$, $\Delta$$s$, and $\Delta$$r$. Furthermore, transient elements in unconstrained images have inconsistent colors across views. Therefore, we additionally introduce the transient color MLP $f_c$ to map the color of transient elements expressed as:
\vspace{-0.3em}
\begin{equation}
(\Delta \mu, \Delta s, \Delta r) = f_d(\gamma(\operatorname{sg}(\mu)), \gamma(t)), \quad \hat{c} = f_c(\gamma(t)),
\end{equation}	where \(\operatorname{sg}(\cdot)\) is the stop-gradient operation, and $\gamma(\cdot)$ denotes the positional encoding. Finally, the transient 3D gaussians is computed as $\mathcal{G}_{d}( \mu_i + \Delta \mu_i,\, s_i + \Delta s_i,\, r_i + \Delta r_i )$ to represent transient elements. Our method differs from Deform3D-GS \cite{yang2024deformable} in one important way: instead of representing dynamic elements that are successively consistent in each image, our primary goal is to efficiently capture ephemeral and transient elements by leveraging deformable transient field showing significant memory efficiency.
\begin{figure*}[!t]
\centering
\includegraphics[width=\textwidth]{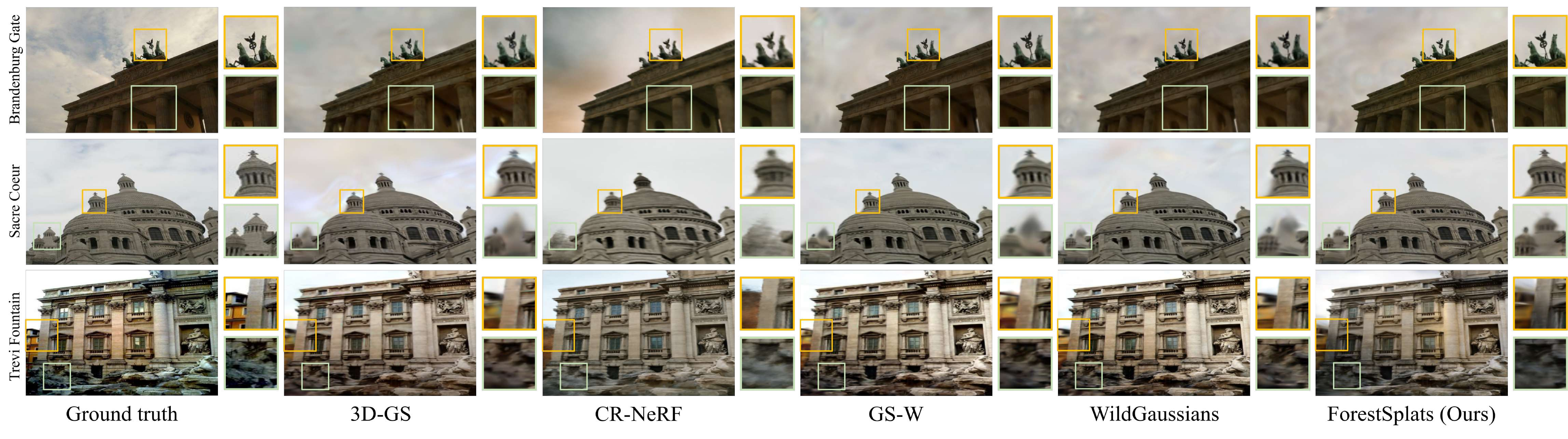}
\vspace{-2.4em}
\caption{Qualitative results from novel-view synthesis on the Photo Tourism dataset. ForestSplats demonstrates geometry consistency and captures fine details compared to existing methods. Yellow and green crop emphasize appearance differences and consistency differences.}
\label{fig:CompPhoto}
\end{figure*}
\begin{figure*}[!t]
\centering
\includegraphics[width=\textwidth]{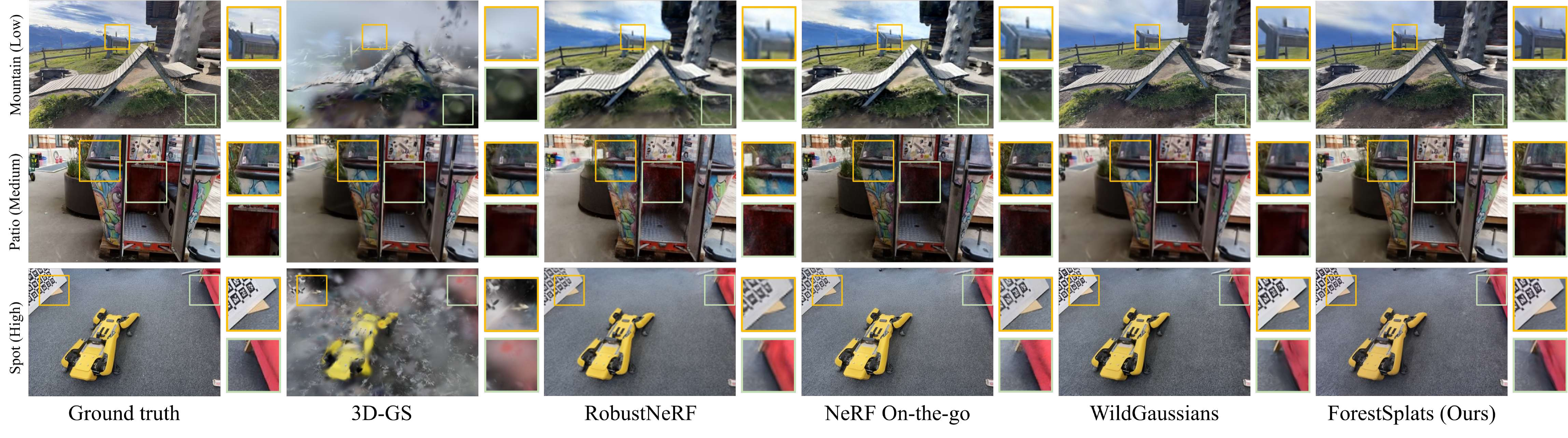}
\vspace{-2.4em}
\caption{Qualitative results from novel-view synthesis on the NeRF On-the-go dataset. ForestSplats effectively removes distractors and reduces artifacts compared to prior methods. Yellow and green crop primarily emphasize artifacts differences and consistency differences.}
\label{fig:CompOnthego}
\end{figure*}
\subsection{Superpixel-aware Mask}
\label{sec:4-2}
\hspace{1em} Existing methods \cite{sabour2023robustnerf, zhang2024gaussian,
kulhanek2024wildgaussians, sabour2024spotlesssplats} primarily address outlier handling using Least Squares techniques or by extracting semantic features from the Vision Foundation Model (VFM). Inspired by these works, we propose a superpixel-aware mask and a multi-stage training scheme that effectively decompose transient distractors from static scenes. Our approach considers photometric errors and superpixels. Following CR-NeRF \cite{yang2023cross}, we obtain the transient mask $\mathcal{M}_{O}$ from Unet $S_{\theta}$ through a learning-based approach, optimizing to minimize the training loss without per-scene optimization. However, directly using the transient mask $\mathcal{M}_{O}$ often struggles with imprecise boundaries of occluders, as shown in Fig. \ref{fig:method-mask}. To tackle this, we leverage the superpixel algorithm \cite{van2015seeds}, which considers color and spatial information. Given an each view $I_{gt}$ partitioned into set of superpixels $\mathcal{S} = \{s_1, ..., s_j\}$, we compute the mask coverage ratio $\rho_{j}$ for each superpixel $s_j$ as follow:{\small\begin{equation}
    \rho_{j} = \frac{|\{p \in s_j \mid \mathcal{M}_{O^{*}}(p) = 1\}|}{|s_{j}|} , \; \mathcal{M}_{O^{*}} = \mathcal{M}_{O} > 0.5,
\end{equation}}where $p$ is pixel position. We employ a threshold to prevent Gaussians from being ambiguously assigned between the static and transient fields. Consequently, the superpixel-aware mask $\mathcal{M}_{S}$ is defined as: 
\vspace{-0.2em}
{\small\begin{equation}
\mathcal{M}_{S}(p) =
\begin{cases}
\max\limits_{p_{} \in s_j} \mathcal{M}_{O^{*}}(p), & \text{if } \rho_j \geq 0.5 \text{ for } p \in s_j \\
\mathcal{M}_{O^{*}}(p), & \text{otherwise}
\end{cases}
\end{equation}}Additionally, to prevent $\mathcal{M}_{O}$ from masking everything, we add a regularization term following CR-NeRF \cite{yang2023cross}.
Furthermore, to mitigate the undesired blending of Gaussians between static and transient fields, we introduce the loss term $\mathcal{L}_{BCE}(\mathcal{M}_{O}, \mathcal{M}_{S})$. Finally, the total loss function is defined as follows:
\vspace{-0.5em}
{\small \begin{align}
\label{eq:l_gs}
\mathcal{L}_{\mathrm{total}}=(1-\lambda) \mathcal{L}_1( \hat{I}, I_{gt}) + 
\lambda \mathcal{L}_{\text{D-SSIM}}(\hat{I},  I_{gt}) \nonumber \\ + \lambda_{0}\mathcal{L}_{BCE}(\mathcal{M}_{O}, \mathcal{M}_{S}) + \lambda_{1}||\mathcal{M}_{O}||^{2},\end{align}}where $\lambda_{0}$ is $3e-2$, and $\lambda_{1}$ is $5e-4$. We observe that initially incorporating the transient mask during training leads to an ambiguous separation between the static and transient fields. To this end, we propose a multi-stage training scheme to resolve the ambiguity between the static and transient fields. Initially, we train the static field to represent the entire image as follows: \begin{equation}{\small
    \mathcal{L}_{\mathrm{init}} = 
    (1-\lambda) \mathcal{L}_1(\hat{I}_{s}, I_{gt}) + 
    \lambda \mathcal{L}_{\text{D-SSIM}}(\hat{I}_{s}, I_{gt}).}
\end{equation}After enough training time, we train the Unet $S_{\theta}$ and the transient field to capture the transient elements through the following learning process:
\begin{equation}{\fontsize{8.4pt}{6.5pt}
    \mathcal{L}_{mid} = \mathcal{L}_{total} + \mathcal{L}_1(\hat{I}_{d} \odot \bar{\mathcal{M}}_{S}, I_{gt} \odot \bar{\mathcal{M}}_{S}), }
\end{equation}where $\bar{\mathcal{M}}_{S}$ is calculated as $\bar{\mathcal{M}}_{S} = 1 - \mathcal{M}_{S}$.  Finally, we jointly train static and transient fields as shown in Eq. \ref{eq:l_gs}. This multi-stage training scheme enhances the static field to capture static scenes while achieving high-quality results.
\begin{figure}[!t]
\centering
\includegraphics[width=\linewidth]{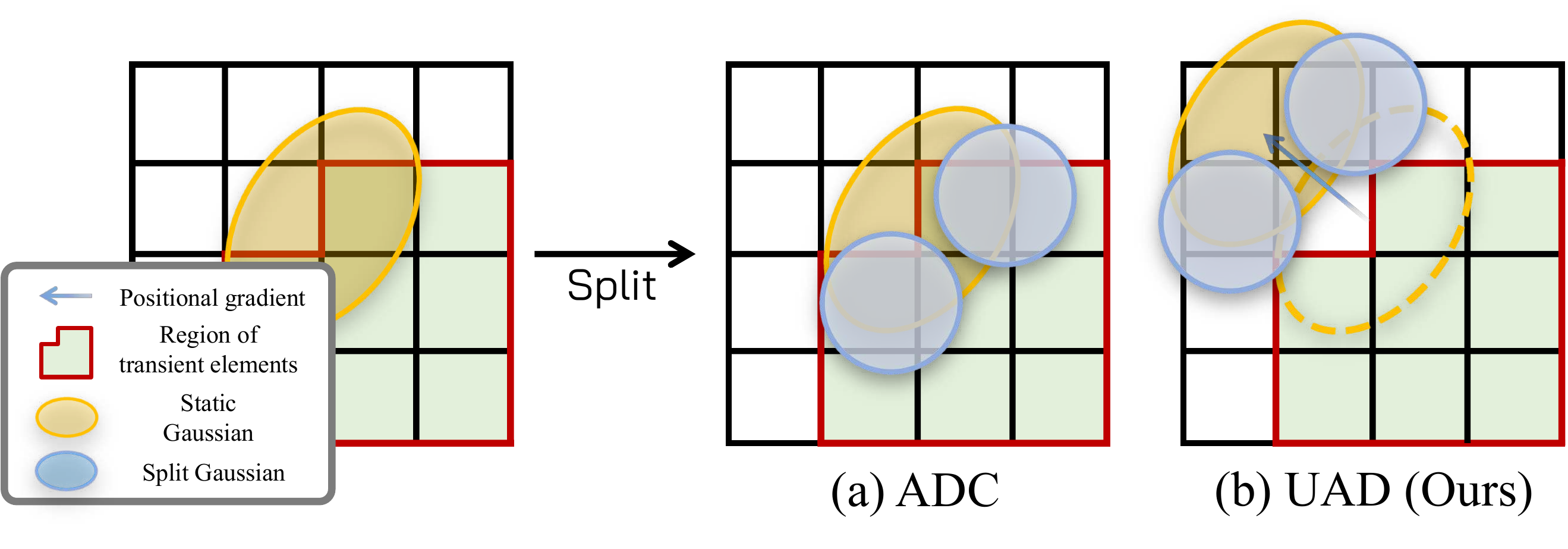}
\vspace{-2em}
\caption{Illustration of ADC and UAD.  (a) The existing adaptive density control (ADC) splits Gaussians by sampling from the PDF. In contrast, (b) Uncertainty-aware densification (UAD) splits Gaussians by considering the direction of positional gradients.}
\label{fig:densification}
\end{figure}
\begin{table*}[!t]
    \centering
    \footnotesize
    \setlength{\tabcolsep}{1.2pt}
    \renewcommand{\arraystretch}{0.94}
    \begin{tabular}{lccccccccccccc}
    \toprule
    \multirow{2}{*}{Method} & \multirow{2}{*}{GPU hrs / FPS} & \multicolumn{3}{c}{Low Occlusion}&\multicolumn{3}{c}{Medium Occlusion}&   \multicolumn{3}{c}{High Occlusion} &   \multicolumn{3}{c}{Average}\\
                              & &  PSNR $\uparrow$ & SSIM $\uparrow$ & LPIPS $\downarrow$ & PSNR $\uparrow$ & SSIM $\uparrow$ & LPIPS $\downarrow$ & PSNR $\uparrow$ & SSIM $\uparrow$ & LPIPS $\downarrow$ & PSNR $\uparrow$ & SSIM $\uparrow$ & LPIPS $\downarrow$ \\
    \midrule
    \midrule
    RobustNeRF \cite{sabour2023robustnerf} & - / $<$1 & 16.60 & 0.407 &0.480 &  21.72 & 0.741 & 0.248 & 20.60 & 0.602  &0.379 & 19.64 & 0.583  &0.369 \\
    Gaussian Opacity Field \cite{yu2024gaussian} & 0.41 / 43 & 20.54 & 0.662 & \underline{0.178} & 19.39 & 0.719 &0.231 &17.81& 0.578 &0.430& 19.24& 0.656  &0.280 \\
    3D-GS \cite{kerbl20233d} & 0.35 / 116 & 19.68 & 0.649 &0.199 & 19.19 & 0.709 &0.220 &19.03& 0.649 &0.340& 19.30& 0.669  &0.253 \\
    Mip-Splatting \cite{yu2024mip} & 0.18 / 82 &  20.15 & 0.661 &0.194 & 19.12 & 0.719 &0.221 &18.10& 0.664 &0.333& 19.12& 0.681  &0.249 \\
    GS-W \cite{zhang2024gaussian} & 1.03 / 105 &  18.67 & 0.595 &0.288 & 21.50 & 0.783 &0.152 &18.52& 0.644 &0.335& 19.56& 0.674  &0.258 \\
    NeRF On-the-go \cite{ren2024nerf}& 43 / $<$1 &  20.63 & 0.661 & 0.191 & 22.31 & 0.780 & 0.130 & 22.19 & 0.753 & 0.169 & 21.71& 0.731  &0.163\\
    WildGaussians \cite{kulhanek2024wildgaussians}& 1.56 / 94 & 20.62 & 0.658 & 0.235 & 22.80 & 0.811 & \textbf{0.092} & 23.03 & 0.771 & 0.172 & 22.15& 0.756  &0.167 \\
    SpotLessSplats$^*$ \cite{sabour2024spotlesssplats}& 1.87 / 81 &  20.01 & 0.596 & 0.276 & 22.79	& 0.816 & 0.161 &	21.91 &	0.710	&0.222	& 21.57	&0.707&	0.219 \\
    DeSplat \cite{wang2024desplat}& 0.84 / 106 & 19.93 & 0.695 & \textbf{0.170} & 23.47& \underline{0.845} & \underline{0.100} & \underline{24.33}& \textbf{0.870}& \textbf{0.105} & 22.58& \underline{0.803}  &\textbf{0.125} \\
    HybridGS \cite{lin2024hybridgs}& 0.98 / 112 & \underline{21.42} & \underline{0.684} & 0.268 & \underline{23.51} & 0.830 & 0.160 & 23.05 & 0.768 & 0.204 & \underline{22.66} & 0.761  &0.211\\
    ForestSplats (Ours) & 1.13  / 117 & 
    \textbf{21.89} & \textbf{0.738} &	0.216 &	\textbf{23.96} &	\textbf{0.870} &	0.135	& \textbf{24.34}	 & \underline{0.815}  &	\underline{0.119} &	\textbf{23.40}	& \textbf{0.807}& \underline{0.156} \\
    \bottomrule
    \end{tabular}
    \vspace{-1em}
    \caption{Quantitative result on NeRF On-the-go dataset. The bold and underlined numbers indicate the best and second-best results. Our method shows competitive performance compared to existing methods. More results are reported in the supplementary material.}
    \label{tab:OnthegoSota}
\end{table*}

\begin{table}[!t]
    \centering
    \footnotesize
    \setlength{\tabcolsep}{5pt} 
    \renewcommand{\arraystretch}{0.94}
    \begin{tabular}{lcccc}
    \toprule
    \multirow{2}{*}{Method}  &\multicolumn{4}{c}{Photo Tourism} \\
      & GPU hrs / FPS & PSNR $\uparrow$ & SSIM $\uparrow$ & LPIPS $\downarrow$ \\
    \midrule
    \midrule
    3D-GS \cite{kerbl20233d}         &  2.2 / 57  & 18.04 & 0.814 & 0.183 \\
    NeRF-W \cite{martin2021nerf}      &  164 / $<$1    & 20.78 & 0.799 & 0.208 \\
    Ha-NeRF \cite{mildenhall2021nerf}  &  452 / $<$1    & 21.41 & 0.793 & 0.178 \\
    DeSplat \cite{wang2024desplat}    &  -  & 22.83 & 0.854 & 0.182 \\
    CR-NeRF \cite{yang2023cross}   &   101 / 0.02  & 23.36 & 0.812 & 0.155 \\
    RobustNeRF \cite{sabour2023robustnerf} & - / $<$1    & 23.44 & 0.853 & 0.134 \\
    IE-NeRF \cite{wang2024ie}   &  - / $<$ 1  & 22.15 & 0.826 &0.181 \\
    SWAG \cite{dahmani2024swag}   &  0.8 / 15  & 23.53 & 0.868 & 0.177 \\
    WildGaussians \cite{kulhanek2024wildgaussians} & 11.5 / 86 & 24.65 & 0.851 & 0.179 \\
    NexusSplats \cite{tang2024nexussplats} & 6.81 / - & 24.95 & 0.849 & 0.185 \\
    GS-W \cite{zhang2024gaussian} & 1.2 / 51       & 24.70 & 0.865 & \underline{0.124} \\
    Wild-GS \cite{xu2024wild} &  0.52 / 227      & \textbf{26.36} & \textbf{0.873} & \textbf{0.128} \\
    ForestSplats (Ours) & 7.9 / 122  & \underline{25.36} &	\underline{0.871} & 0.151  \\
    \bottomrule
    \end{tabular}
    \vspace{-1em}
    \caption{Quantitative average result on the Photo Tourism dataset. The bold and underlined indicate the best and second-best results.}
    \label{tab:PhotoSota}
\end{table}

\subsection{Uncertainty-aware Densification}
\label{sec:4-3}
\hspace{1em} During optimization in joint training with densification, we observe that static field $\mathcal{G}_{s}$ often generates Gaussians near the boundaries of distractors due to random sampling from the probability density function (PDF), as illustrated in Fig.\hyperref[fig:densification]{~\ref{fig:densification}-(a)}. To address this, we propose a simple yet effective uncertainty-aware densification, which introduces uncertainty parameters $l^{(u)}$ for each static Gaussian and considers the positional gradient and the number of pixels covered by the Gaussian. After the initial training, the uncertainty parameters $l^{(u)}$ are optimized guided by superpixel-aware mask $\mathcal{M}_{S}$ as follow:
{\fontsize{8pt}{2pt}\selectfont
\begin{align}
\mathcal{L}_{\text{t}} = -\frac{1}{H \times W} 
\sum_{i=1}^{H \times W} & \bigg[ 
\mathcal{M}_{S,i}\log(\hat{\mathcal{M}}_{l^{(u)},i}) \nonumber \\[-0.9ex]
& + (1 - \mathcal{M}_{S,i}) \log(1 - \hat{\mathcal{M}}_{l^{(u)},i}) 
\bigg], \end{align}}where $\hat{\mathcal{M}}_{l^{(u)}}$ is calculated as $(\sum_{k = 1}^{K}l_{k}^{(u)}) \odot \mathcal{M}_{S}$. According to existing methods \cite{zhang2024pixel,du2024mvgs}, large Gaussians often cover a small area depending on the viewpoint. Inspired by this, we improve the splitting process by shifting static Gaussians along their positional gradient $\mathbf{g}_{i}$, weighted by the number of pixels covered by each Gaussian, computed as $\mathbf{g}_{i} \times( \frac{| \{ p \mid p \in \mathcal{G}_i \} |}{H \times W} )$. Subsequently, we apply random sampling from the PDF to avoid splitting the static Gaussians lying in a region of transient elements as shown in Fig.\hyperref[fig:densification]{~\ref{fig:densification}-(b)}. For above the threshold, similar to 3D-GS-based methods, we remove the static Gaussians with high uncertainty to ensure high-quality and consistent rendering results.
\section{Experiments}
\footnotetext[1]{$^*$The results are reproduced from the official code.}
\paragraph{Datasets, metrics, and baseline.} 
Following existing methods \cite{kulhanek2024wildgaussians}, we evaluate our method on two datasets: the NeRF On-the-go dataset \cite{ren2024nerf} and the Photo Tourism dataset \cite{snavely2006photo}. The NeRF On-the-go dataset \cite{ren2024nerf} contains twelve scenes, casually captured in indoor and outdoor settings, with three varying levels of occlusion (ranging from 5\% to 30\%). For a fair comparison, we utilize six scenes and follow the version of the dataset where all images were undistorted as \cite{kulhanek2024wildgaussians}. The Photo Tourism dataset \cite{snavely2006photo} includes several scenes of well-known monuments. Each scene consists of a collection of user-uploaded unconstrained images captured at different dates, times of day, and exposure levels, covering a wide range of lighting conditions. In our experiments, we utilize three scenes: Brandenburg Gate, Sacre Coeur, and Trevi Fountain. We evaluate our method without including occluders in the image, following previous methods \cite{yang2023cross,kulhanek2024wildgaussians, xu2024wild}. To evaluate our method, we use PSNR, SSIM \cite{wang2004image}, and LPIPS \cite{zhang2018unreasonable} as metrics. \begin{table}[!t]
    \renewcommand{\arraystretch}{0.9}
    \setlength{\tabcolsep}{1.96pt} 
    \scriptsize
    \centering
    \begin{tabular}{cccccc}
    \toprule
        {\scriptsize  Scene                      }     & {\scriptsize Method  }                 & {\scriptsize Memory  }          & {\scriptsize Scene   }                  & {\scriptsize Method    }                &{\scriptsize Memory }\\ \midrule \midrule
\multirow{3}{*}{{\scriptsize\shortstack{Patio\\high}}}             &   Desplat \cite{wang2024desplat}                &  49.96               & \multirow{3}{*}{\scriptsize Corner} & \scriptsize  Desplat \cite{wang2024desplat}   &  31.51     \\
                                        &   HybridGS  \cite{lin2024hybridgs}            &  76.22                 &                           &     HybridGS \cite{lin2024hybridgs}          &   48.07     \\
                                        &   ForestSplats (Ours)     &  3.28           &                             &   ForestSplats (Ours)      &     2.27    \\\midrule \midrule
\multirow{3}{*}{\scriptsize\shortstack{Brandenburg\\Gate}}         &   Desplat   \cite{wang2024desplat}             &      	 304.47                   & \multirow{3}{*}{\scriptsize\shortstack{Sacre\\Coeur}} &   Desplat \cite{wang2024desplat}    & 260.61       \\
                                        &  HybridGS \cite{lin2024hybridgs}              &    464.43                 &                              &   HybridGS \cite{lin2024hybridgs}     &  397.33     \\
                                        &  ForestSplats (Ours)    &      4.41               &                          &    ForestSplats (Ours)    &    3.39     \\
    \bottomrule
    \end{tabular}
    \vspace{-1.2em}
    \caption{Ablation of computational efficiency. We report memory usage (MB) for initial the number of transient Gaussians.}
    \label{tab:ablMem}
\end{table} \begin{figure}[!t]
    \centering
    \includegraphics[width=\linewidth]{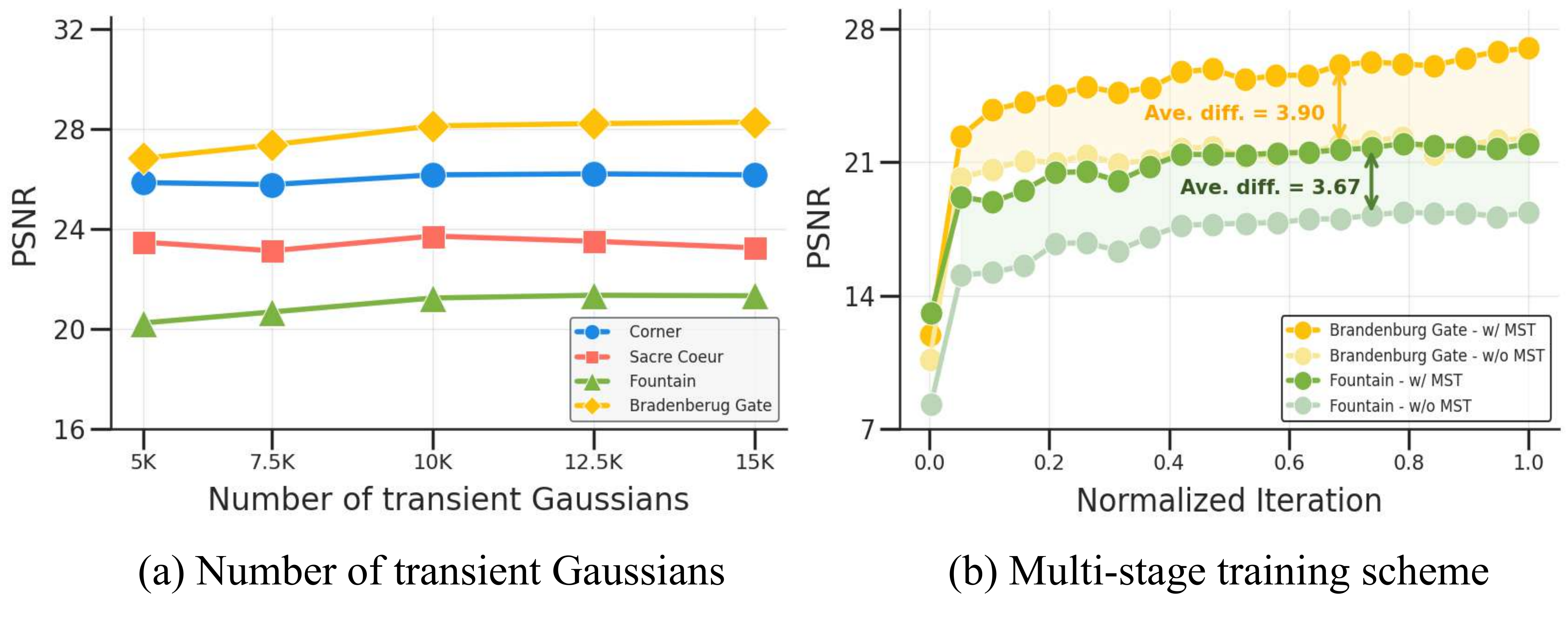}\vspace{-1em}
    \caption{Ablation of the number of transient Gaussians and the multi-stage training scheme. (a) Evaluation of results across different numbers of transient Gaussians. (b) Comparison of training PSNR with and without a multi-stage training scheme.}
    
    \label{fig:tr-gsstrategy}
\end{figure}
\vspace{0.5\baselineskip}
\newline \textbf{Implementation details.} We develop our ForestSplats method based on WildGaussians \cite{kulhanek2024wildgaussians}. We follow the hyperparameter settings for the default setting and train using the Adam \cite{kingma2014adam} optimizer without weight decay. Specifically, we conducted the improved ADC solely for a static field. Furthermore, we extended the implementation by adapting the Pixel-GS \cite{zhang2024pixel} to calculate the number of per-pixel Gaussians using a custom CUDA kernel. For transient elements, we adopt the transient field initialized by randomly selecting 10K points from those obtained through SfM. In the initial training phase, we conduct 150K steps of training solely on the static field. Subsequently, we proceed to joint training, where both static and transient fields are optimized until 200K iterations. Finally, we focus on training the static field to capture the static scene. All experiments are performed on an NVIDIA RTX 24GB GPU using NerfBaselines \cite{kulhanek2024nerfbaselines} as our evaluation framework for a fair comparison.

\section{Evaluation}
\paragraph{Comparison on the Photo Tourism Dataset.} As shown in Fig. \ref{fig:CompPhoto} and Tab. \ref{tab:PhotoSota}, our method shows favorable performance compared to prior methods. In Fig. \ref{fig:CompPhoto}, we observe that CR-NeRF \cite{yang2023cross}, 3D-GS \cite{kerbl20233d}, and GS-W \cite{zhang2024gaussian} frequently fail to capture fine details in scenarios with appearance variations and struggle to handle transient elements. Unlike existing methods, WildGaussians \cite{kulhanek2024wildgaussians} and Wild-GS \cite{xu2024wild} leverage semantic features from VFM \cite{yang2024depth, oquab2023dinov2} to handle the transient elements. In contrast, our ForestSplats effectively decomposes transient elements from 2D scenes by leveraging transient field and a superpixel-aware mask without VFM, as shown in Tab. \ref{tab:PhotoSota}. More details are in the supplements.
\begin{table}[!t]
    \scriptsize
    \renewcommand{\arraystretch}{0.87}
    \centering
    \resizebox{1.0\columnwidth}{!}{
    \begin{tabular}{ccccc}
    \toprule
    Scene   & Transient Field   & PSNR $\uparrow$  & SSIM $\uparrow$  & LPIPS $\downarrow$  \\
\midrule
\midrule
                \multirow{2}{*}{Corner} & w/o deformable   & 24.52  & 0.855  & 0.153            \\
                 &       w/ deformable   & 26.17	 & 0.891 & 0.136    \\ 
\midrule
                \multirow{2}{*}{\scriptsize\shortstack{Brandenburg\\Gate}}  & w/o deformable    &   26.83  & 0.925   & 0.133    \\
                &       w/ deformable    & 28.13  & 0.935  & 0.128  \\
    \bottomrule
    \end{tabular}
    }
    \vspace{-1.1em}
    \caption{Ablation of the with and without Deformable for effect of transient field on the Brandenburg Gate and Corner scenes.}
    \label{tab:staticdeformabl}
\end{table}
\begin{figure}[!t]
    \centering
    \includegraphics[width=\linewidth]{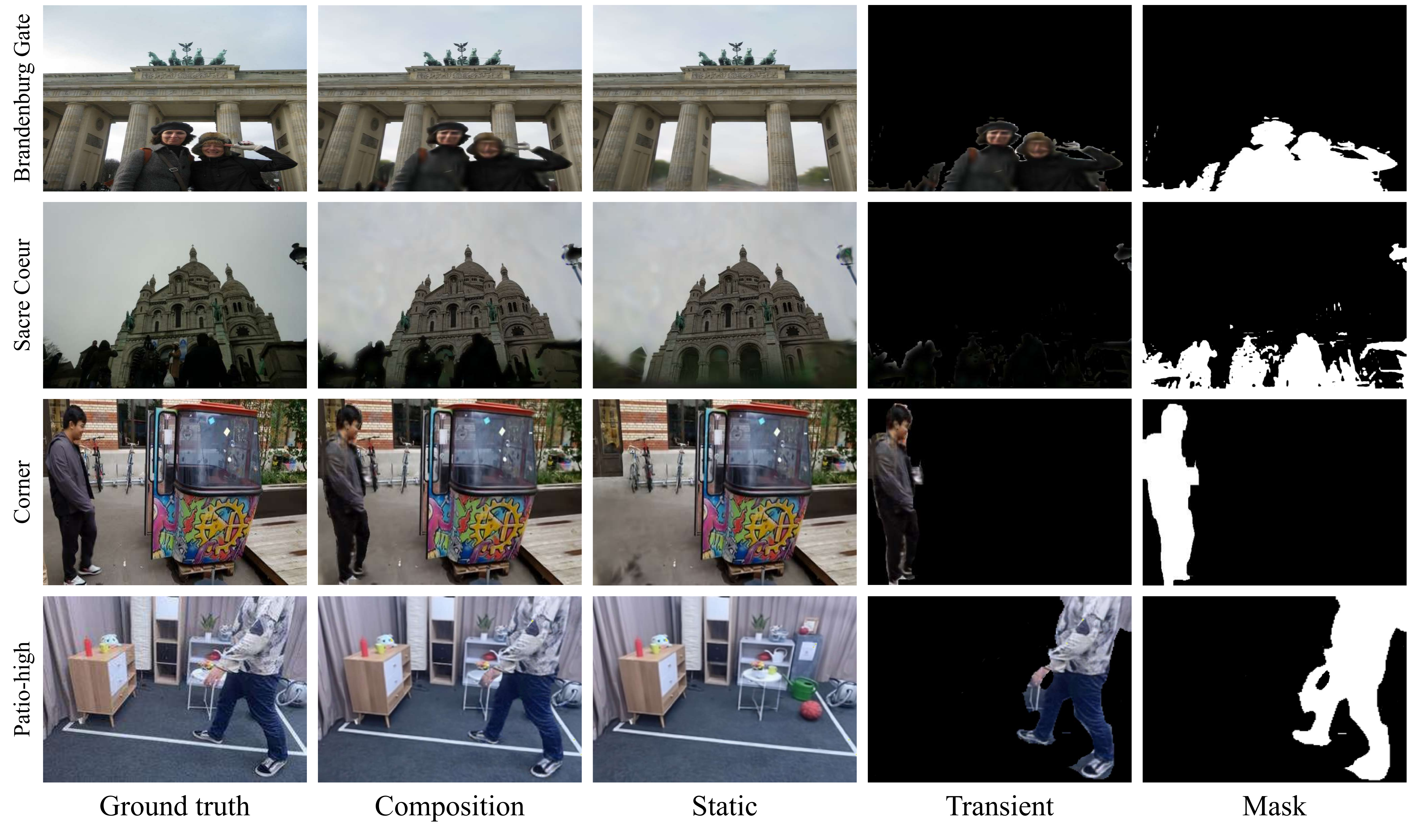}
    \vspace{-2.2em}
    \caption{Ablation of the novel-view synthesis for static and transient fields on the NeRF On-the-go and Photo Tourism datasets.}
    \label{fig:compositionabl}
\end{figure}

\vspace{0.5\baselineskip}
\noindent \textbf{Comparison on the NeRF On-the-go Dataset.} Furthermore, we also show the results for the NeRF On-the-go dataset \cite{ren2024nerf}. Our method achieves competitive performance against existing SOTA methods across various occlusion levels without VFM in Tab. \ref{tab:OnthegoSota}. Compared to RobustNeRF \cite{sabour2023robustnerf}, and NeRF On-the-go \cite{ren2024nerf}, our method exhibits faster training and rendering speeds while mitigating occluders. RobustNeRF \cite{sabour2023robustnerf} considers photometric errors and the small size of the sampled patches to overlook complex structures and high occlusion scenarios. Specifically, WildGaussians \cite{kulhanek2024wildgaussians} and SpotLessSplats \cite{sabour2024spotlesssplats} effectively remove transient elements by considering semantic features. Although leveraging semantic features facilitates handling transient elements, distinguishing semantic features similar to static elements remains challenging. However, our method explicitly decomposes static and transient elements across different levels of occluder scenarios, as shown in Fig. \ref{fig:CompOnthego}. Furthermore, compared to the prior methods without VFM, our method outperforms performance in terms of PSNR and SSIM.

\begin{figure}[!t]
    \centering
    \includegraphics[width=\linewidth]{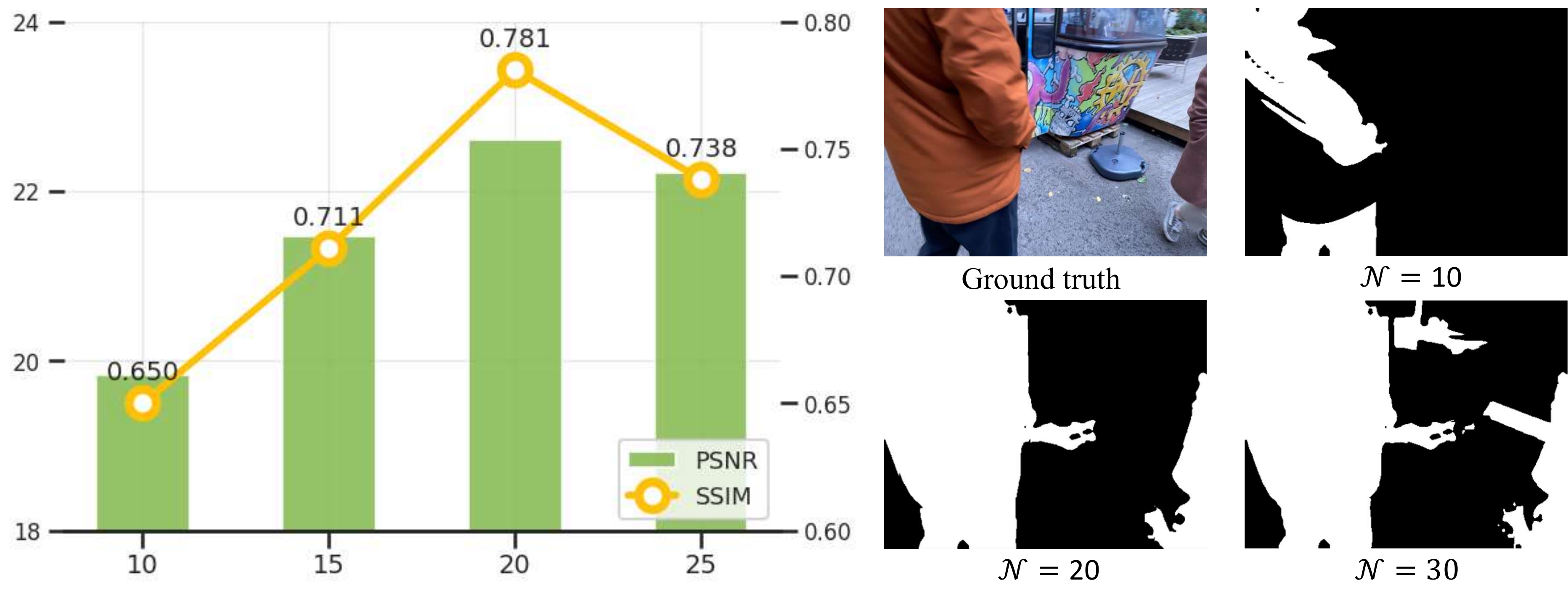}
    \vspace{-2em}
    \caption{Ablation of the number of superpixels on the Patio-high scene. $\mathcal{N}$ denotes the number of superpixels.}
    \label{fig:numsegment}
\end{figure}
\begin{figure}[!t]
  \begin{minipage}[b]{.35\linewidth}
      \subcaptionbox{Comparison of transient mask\label{sub:depth}}{
      \includegraphics[width=1.3\columnwidth, height=3.0cm]{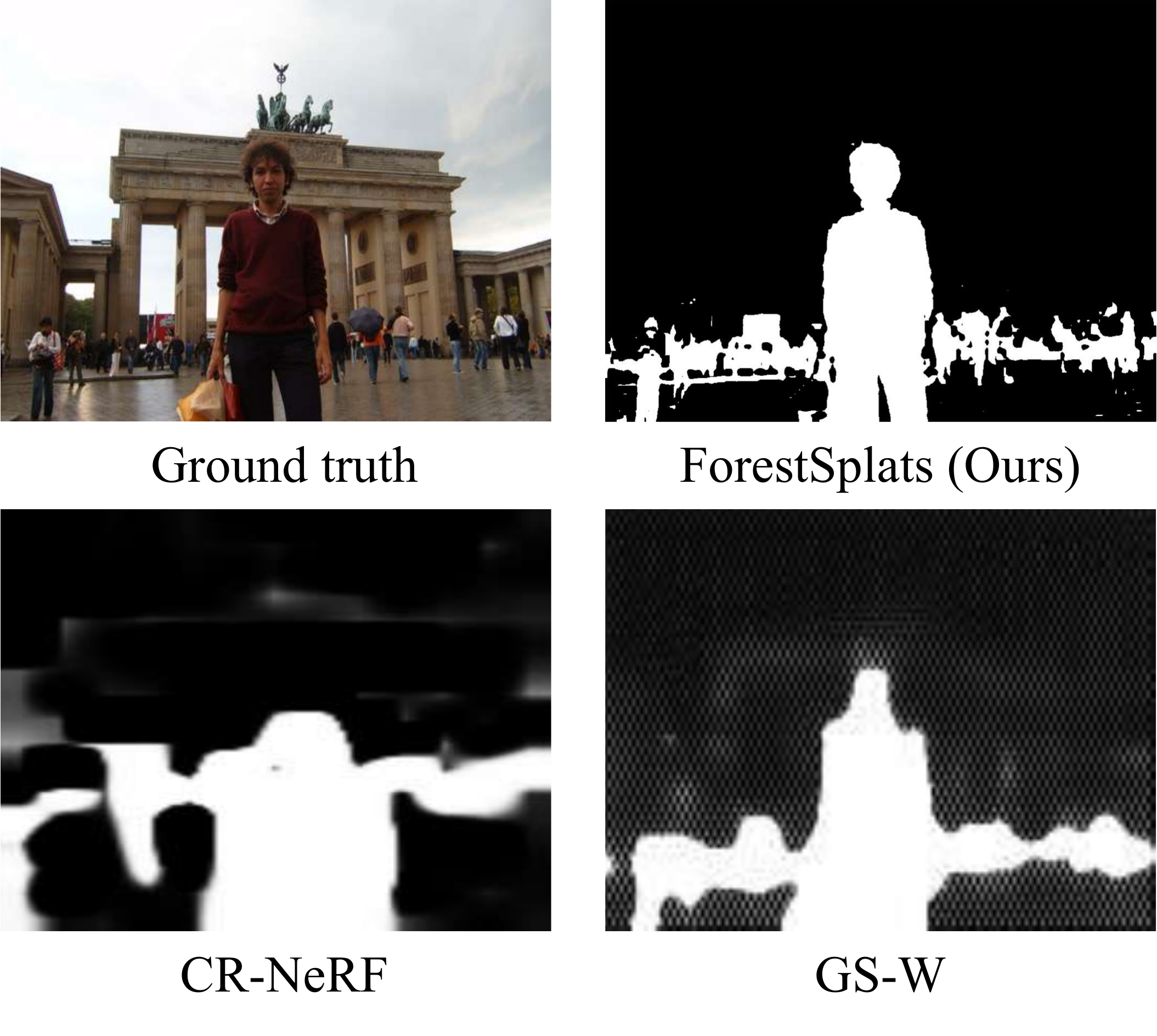} 
      }
    \end{minipage} 
    \hfill
    \begin{minipage}[b]{.78\linewidth}
        \footnotesize
        \centering
        \setlength{\tabcolsep}{0.005pt}
        \renewcommand{\arraystretch}{1.5}
        \subcaptionbox{\footnotesize{Impact of superpixel-aware mask.}\label{sub:module}}{
        \begin{tabular}{lccc}
        \toprule
        \scriptsize{Method} & \; \scriptsize{PSNR $\uparrow$} & \; \scriptsize{SSIM $\uparrow$} & \; \scriptsize{LPIPS $\downarrow$} \\
        \midrule
        \midrule
        \scriptsize{CR-NeRF \cite{yang2023cross}} &26.53 & 0.900 & 0.106 \\
        \scriptsize{\;\;$+$ $\mathcal{M}_{s}$}        & 27.13 & 0.908 & 0.077 \\
        \scriptsize{GS-W  \cite{zhang2024gaussian}} & 27.96 & 0.931 & 0.086 \\
        \scriptsize{\;\;$+$ $\mathcal{M}_{s}$}      & 28.62 & 0.926 & 0.048 \\
        \bottomrule
        \end{tabular}
        }
    \end{minipage}
    \vspace{-0.8em}
\caption{Ablation of the transient mask. (a) Comparison of transient mask methods. (b) Effect of the superpixel-aware mask.}
\label{fig:compareMaskabl}
\end{figure}
\section{Ablation Study}
\paragraph{Efficiency of Deformable Transient Field.} As shown in Tab. \ref{tab:ablMem}, the deformable transient field significantly reduces the memory usage (MB). Specifically in the Sacre Coeur, although transient embedding per image is considered, the deformable transient field achieves over 80\% memory efficiency (MB). Compared to existing methods \cite{wang2024desplat, lin2024hybridgs}, the deformable transient field adjusts the transient Gaussians based on per-scene optimization, showing efficient rendering for transient elements. Furthermore, we also analyze the performance based on the number of transient Gaussians as shown in Fig. \hyperref[fig:tr-gsstrategy]{\ref{fig:tr-gsstrategy}-(a)}. We observe that the number of transient Gaussians primarily aims to occupy the regions with transient elements, which results in a slight effect on the representation of the static field. Moreover, we compare the transient field with and without the deformation MLP, as shown in Tab. \ref{tab:staticdeformabl}. These results indicate that the deformable transient field not only significantly reduces memory usage (MB) but also facilitates scalability, particularly in large-scale scenarios.
\begin{figure}[!t]
    \centering
    \includegraphics[width=\linewidth]{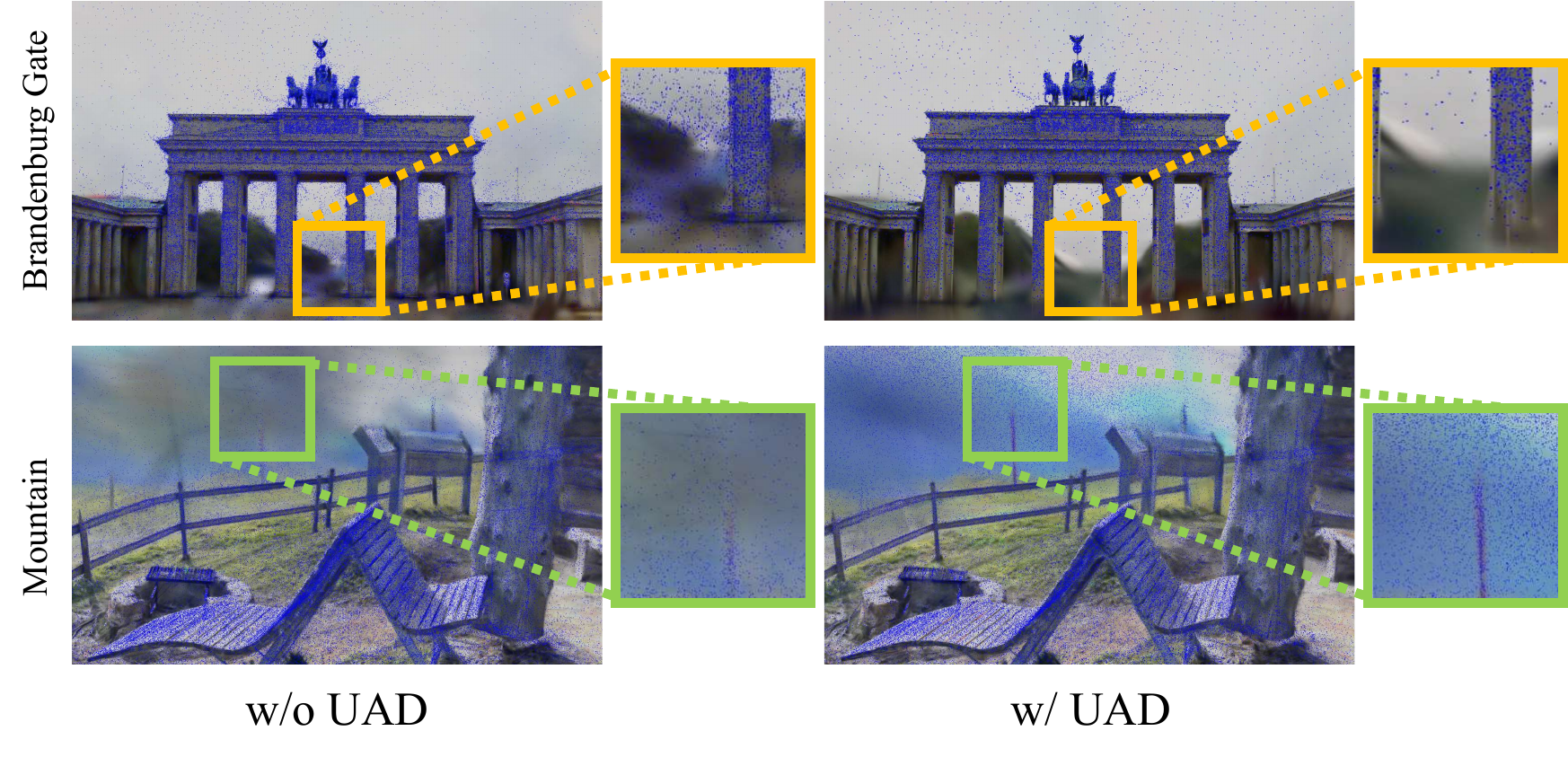}
    \vspace{-2.5em}
    \caption{Ablation of the uncertainty-aware densification (UAD). We visualize centroids of static Gaussians with and without UAD.}
    \label{fig:wouad}
\end{figure}
\begin{figure}[!t]
    \centering
    \includegraphics[width=\linewidth]{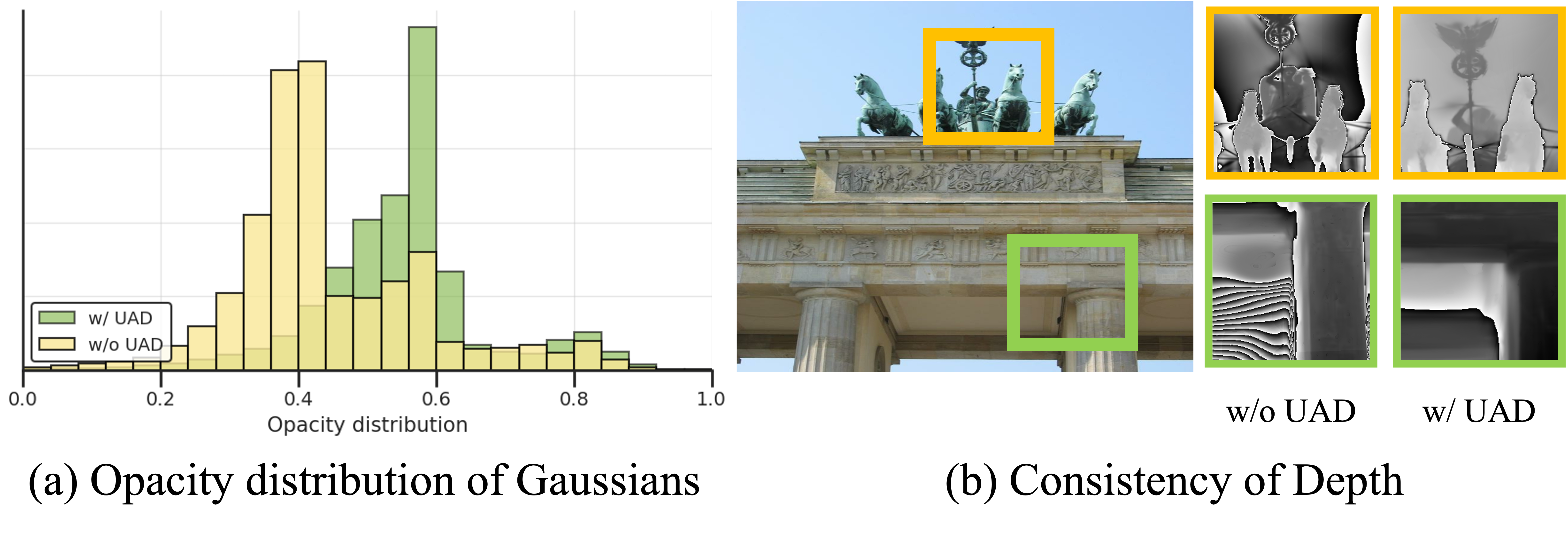}\vspace{-1.2em}
    \caption{Ablation of uncertainty-aware densification (UAD) on opacity distribution and depth consistency. (a) Opacity distribution of Gaussians. (b) Consistency of depth. We visualize the effect of uncertainty-aware densification (UAD) on opacity and depth.}   
    \label{fig:uad-opacity}
\end{figure}

\begin{figure}[!t]
    \centering
    \includegraphics[width=\linewidth]{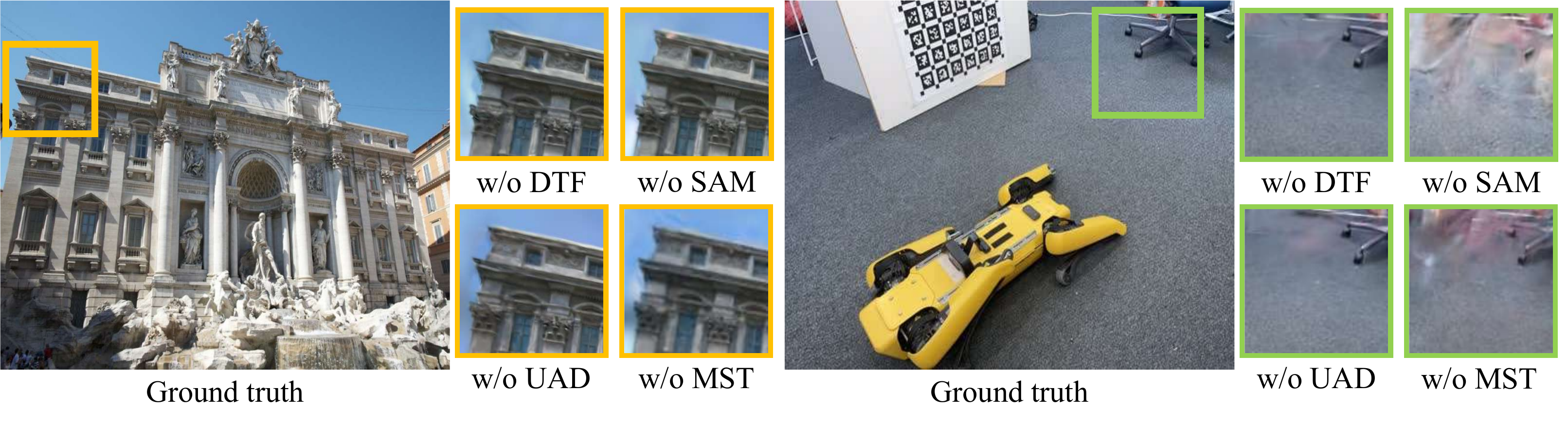}
    \vspace{-2.4em}
    \caption{Ablation study of the qualitative results for each component in ForestSplats. More details are provided in the supplements.}
    \label{fig:compovis}
\end{figure}

\vspace{0.5\baselineskip}
\noindent \textbf{Effectiveness of superpixel-aware mask.} We conducted an ablation study to assess the impact of a superpixel-aware mask in handling transient elements. As shown in Fig. \ref{fig:compositionabl}, our method explicitly decomposes distractors from the static scene and represents transient objects in various scenarios, handling varying levels of occlusion. Furthermore, we analyze the effect of the number of superpixels in Fig. \ref{fig:numsegment}. According to the results, a small number of superpixels has minimal effect on the transient mask. Meanwhile, a large number of superpixels may lead to excessive masking of irrelevant areas. Thus, we set the number of segments to 20 as an optimal number. Moreover, we compare the quality of the transient mask in Fig. \hyperref[fig:compareMaskabl]{\ref{fig:compareMaskabl}-(a)}. Existing methods \cite{yang2023cross, zhang2024gaussian} rely on photometric errors. In contrast, superpixel-aware masks leverage superpixels, which consider both color and spatial information, to enhance mask quality without additional memory. Furthermore, the superpixel-aware mask serves as a plug-and-play module that combines with prior methods, showing improvement, as shown in Fig. \hyperref[fig:compareMaskabl]{\ref{fig:compareMaskabl}-(b)}. In addition, we demonstrate a multi-stage training scheme that consistently improves performance in Fig. \hyperref[fig:tr-gsstrategy]{\ref{fig:tr-gsstrategy}-(b)}.
\vspace{0.5\baselineskip}
\newline \textbf{Discussion on uncertainty-aware densification.} We use uncertainty-aware densification (UAD) only in the static field to avoid the generation of static Gaussians at the boundaries of distractors during optimization. To analyze this, we examine an ablation study with and without UAD to see its effect on novel-view synthesis. As shown in Fig. \ref{fig:wouad}, we observe that the centroids of static Gaussians are primarily positioned on static elements, mitigating the artifacts in novel view synthesis. In addition to, We analyze the effect of UAD on opacity and depth in Fig. \ref{fig:uad-opacity}. Our results show that UAD yields Gaussians exhibiting higher opacity values and improved depth consistency.
\vspace{0.5\baselineskip}
\newline \textbf{Analysis of each component.} For further analysis, we evaluate each component of our method to demonstrate the effectiveness: Deformable transient field (DTF), Superpixel-aware mask (SAM), Multi-stage training scheme (MST), and Uncertainty-aware densification (UAD) as shown in Tab. \ref{tab:component}. The results indicate that a superpixel-aware mask, which considers the photometric errors and the superpixels, combined with a multi-stage training scheme, consistently improves performance by effectively removing transient elements in scenes. Furthermore, we show qualitative results on the Brandenburg
Gate and Corner scenes in Fig. \ref{fig:compovis}.
\begin{table}[!t]
    \centering
    \footnotesize
    \setlength{\tabcolsep}{7pt} 
    \renewcommand{\arraystretch}{0.9}
    \resizebox{1.0\columnwidth}{!}{
    \begin{tabular}{lcccccc}
        \toprule
        \multicolumn{4}{l}{Methods} & PSNR $\uparrow$ & SSIM $\uparrow$ & LPIPS $\downarrow$ \\
        \midrule
        \midrule
        \multicolumn{4}{l}{Baselines \cite{wang2024desplat}} & 20.14 & 0.868 & 0.178 \\
        \multicolumn{4}{l}{+ Deformable transient Field}  & 22.38 &  0.861 &  0.175 \\
        \multicolumn{4}{l}{+ Superpixel-aware Mask} & 22.75 & 0.869 & 0.162 \\
        \multicolumn{4}{l}{\;\;\;\; + Multi-stage training scheme} &  23.20 & 0.861 & 0.131\\
        \multicolumn{4}{l}{+ Uncertainty-aware Densification} & 23.31 & 0.867 & 0.146 \\
        \multicolumn{4}{l}{\;\;\;\; + Pixel-aware gradient (Full)} & 23.84 & 0.876 & 0.123 \\
        \bottomrule
    \end{tabular}}
    \vspace{-1.3em}
    \caption{Ablation of the effectiveness of each component in ForestSplats. We report all quantitative results on the Sacre Coeur scene.}
    \label{tab:component}
\end{table}
\section{Limitation and Conclusion}
\paragraph{Limitation.} Although the transient field does not exhibit high quality due to deformable Gaussians, as shown in Fig. \ref{fig:compositionabl}, the deformable transient field effectively captures transient elements while showing memory efficiency. Another limitation of our forestsplats is its difficulty in generalizing the number of superpixels due to the varying levels of occlusion in datasets, as depicted in Fig. \ref{fig:numsegment}. 
\vspace{0.5\baselineskip}
\newline \textbf{Conclusion.} In this paper, we introduce ForestSplats, a novel framework that leverages a deformable transient field and a superpixel-aware mask to effectively remove distractors from static scenes. Furthermore, we propose a simple yet effective uncertainty-aware densification to enhance rendering quality and avoid generating static Gaussians on the boundaries of distractors during optimization. Extensive experiments on several datasets demonstrate that our method shows state-of-the-art performance compared to existing methods without a pre-trained model and shows significant memory efficiency to represent transient elements.
{
    \small
    \bibliographystyle{ieeenat_fullname}
    \bibliography{main}
}

\end{document}